\documentclass[10pt,twocolumn,letterpaper]{article}

\usepackage{cvpr}              % To produce the CAMERA-READY version
\usepackage{amssymb}
\usepackage{latexsym}
\usepackage{booktabs}
\usepackage{url}
\usepackage{xcolor}
\usepackage[colorlinks]{hyperref}
\usepackage{mathtools}
\usepackage{bbding}
\usepackage{soul}
\usepackage{bibunits}
\usepackage{bm}
\usepackage{natbib}
\usepackage{array}
\usepackage{pdflscape}
\usepackage{makecell}
\usepackage{framed,multirow}
\usepackage{threeparttable}
\usepackage{amssymb}% http://ctan.org/pkg/amssymb
\usepackage{pifont}% http://ctan.org/pkg/pifont
\usepackage{algorithm}
\usepackage{layouts}
\usepackage{listings}
\usepackage{colortbl}
\usepackage{pythonhighlight}
\makeatletter
\newcommand\footnoteref[1]{\protected@xdef\@thefnmark{\ref{#1}}\@footnotemark}
\makeatother

\newcolumntype{P}[1]{>{\centering\arraybackslash}p{#1}}
\newlength\savewidth
\def\arrvline{\hfil\kern\arraycolsep\vline\kern-\arraycolsep\hfilneg}

\definecolor{Highlight}{HTML}{39b54a}  % green

\definecolor{iblue}{rgb}{0.06, 0.75, 1.0}
\definecolor{igray}{rgb}{0.00, 0.00, 0.00}
\definecolor{ired}{rgb}{0.8588, 0.2666, 0.2156}

\usepackage{aligned-overset}
\usepackage{cases}

\definecolor{newcolor}{rgb}{.8,.349,.1}

\definecolor{cvprblue}{rgb}{0.21,0.49,0.74}
\def\paperID{*****} % *** Enter the Paper ID here
\def\confName{CVPR}
\def\confYear{2026}

\title{Are Video Models Emerging as Zero-Shot Learners and Reasoners \\in Medical Imaging?}
\author{
Yuxiang Lai$^{1}$\thanks{Authors contributed equally.} \quad
Jike Zhong$^{2}$\footnotemark[1] \quad
Ming Li$^{3}$ \quad
Yuheng Li$^{4}$ \quad
Xiaofeng Yang$^{1,5,6}$\thanks{Corresponding author: \texttt{xiaofeng.yang@emory.edu}}\\[4pt]
$^{1}$Department of Computer Science, Emory University\\
$^{2}$Department of Computer Science, University of Southern California\\
$^{3}$Department of Computer Science, University of Maryland\\
$^{4}$Department of Biomedical Engineering, Georgia Institute of Technology\\
$^{5}$Department of Radiation Oncology and Winship Cancer Institute, Emory University\\[2pt]
}

\begin{document}

\maketitle

% \begingroup
% \renewcommand\thefootnote{}\footnotetext{This research is supported in part by the National Institutes of Health under Award Numbers R01EB032680, R01CA272991, R01DE033512, and U54CA274513.}
% \addtocounter{footnote}{-1}
% \endgroup

\begin{abstract}
Recent advances in large generative models have shown that simple autoregressive formulations, when scaled appropriately, can exhibit strong zero-shot generalization across domains. Motivated by this trend, we investigate whether autoregressive video modeling principles can be directly applied to medical imaging tasks, despite the model never being trained on medical data. Specifically, we evaluate a large vision model (LVM) in a zero-shot setting across four representative tasks: organ segmentation, denoising, super-resolution, and motion prediction.
Remarkably, even without domain-specific fine-tuning, the LVM can delineate anatomical structures in CT scans and achieve competitive performance on segmentation, denoising, and super-resolution. Most notably, in radiotherapy motion prediction, the model forecasts future 3D CT phases directly from prior phases of a 4D CT scan, producing anatomically consistent predictions that capture patient-specific respiratory dynamics with realistic temporal coherence.
We evaluate the LVM on 4D CT data from 122 patients, totaling over 1,820 3D CT volumes. Despite no prior exposure to medical data, the model achieves strong performance across all tasks and surpasses specialized DVF-based and generative baselines in motion prediction, achieving state-of-the-art spatial accuracy. These findings reveal the emergence of zero-shot capabilities in medical video modeling and highlight the potential of general-purpose video models to serve as unified learners and reasoners—laying the groundwork for future medical foundation models built on video models.
\end{abstract}

\vspace{-15pt}

\section{Introduction}\label{sec:introduction}
\vspace{-5pt}

\begin{figure*}[t]
	\centering
\includegraphics[width=0.8\linewidth]{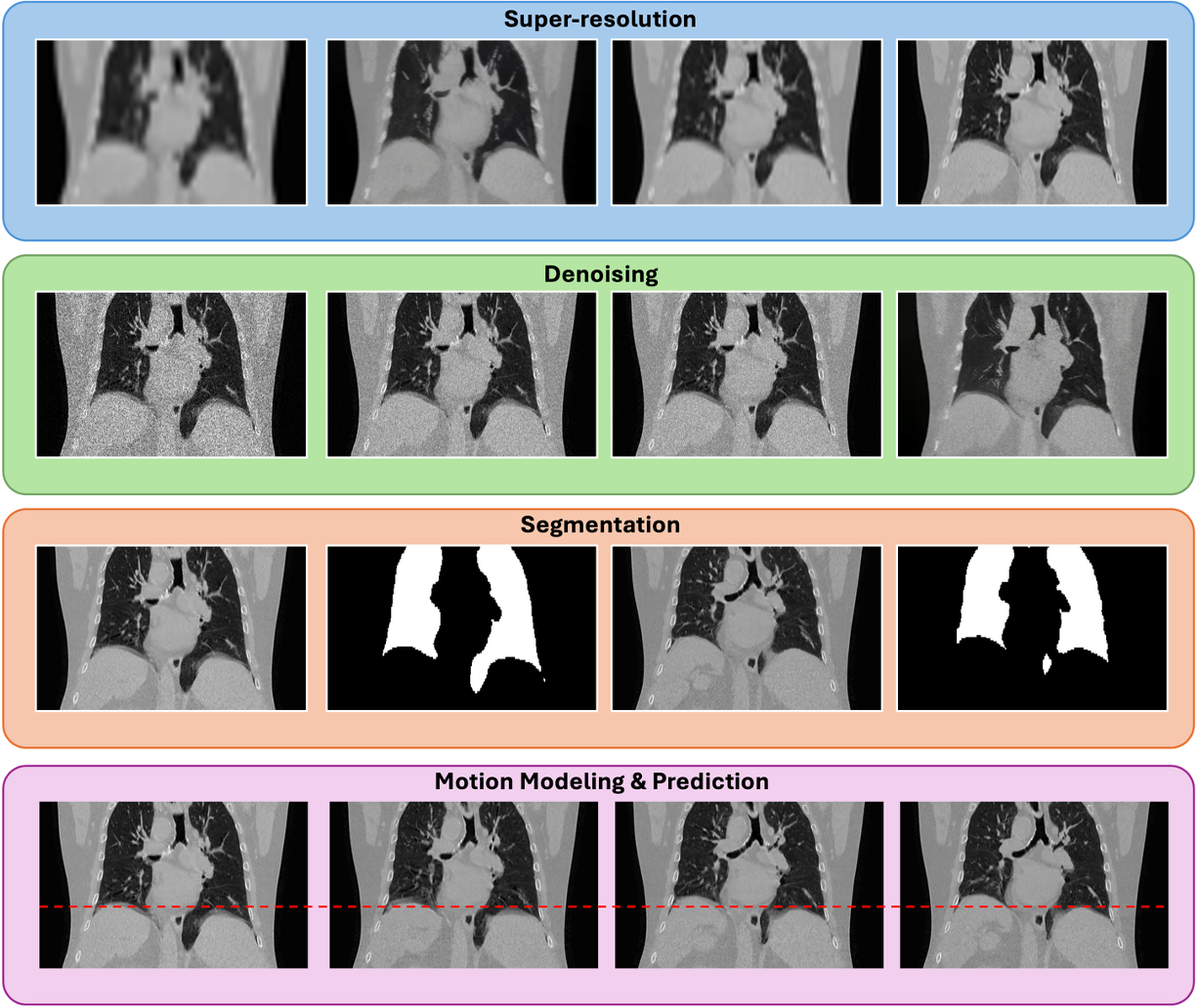}
\caption{Zero-shot learning and reasoning examples of the video model in medical imaging.
From low-level perceptual restoration (super-resolution, denoising) to high-level understanding tasks (segmentation, motion modeling, and prediction), 
the video model can perform a range of medical imaging tasks directly from CT sequences without task-specific training. 
The examples highlight the potential to further advance video models toward becoming foundational vision models for medical imaging. 
.}

    \label{fig_example_tasks}
\end{figure*}

%\linenumbers

Large Language Models (LLMs) and Vision-Language Models (VLMs) have fundamentally reshaped how complex problems are approached. Traditionally, medical imaging tasks required carefully engineered pipelines—for example, using separate models for segmentation, measurement, and motion prediction. In contrast, recent medical foundation models such as MedGemma~\cite{sellergren2025medgemma} demonstrate that unified frameworks can perform multiple tasks within a single system, reducing reliance on task-specific components. Similar to their general-domain counterparts, medical LLMs and VLMs are increasingly capable of tackling novel problems through few-shot in-context learning~\cite{brown2020language,radford2019language} and even zero-shot learning, where task instructions replace fine-tuning or custom prediction heads.  

Building on this trend, \citet{wiedemer2025video} argue that vision models are following a similar trajectory to NLP—shifting from task-specific architectures toward unified foundation models. This suggests that large-scale video models may represent the next generation of general-purpose visual backbones capable of temporal reasoning and multi-task generalization.  

However, general-purpose medical vision models remain largely underexplored. While task-specific architectures such as U-Net~\cite{isensee2021nnu, chen2021transunet, ronneberger2015u} have achieved remarkable success in segmentation, no existing framework unifies medical imaging tasks such as segmentation, registration, and motion prediction under a single paradigm. In medicine, where tasks remain highly specialized and pipelines fragmented, the potential of video models to generalize across tasks without retraining is particularly promising for building scalable and adaptive medical AI systems.

In this work, we adopt Large Vision Model (LVM)~\cite{bai2024sequential} as our baseline video modeling backbone. LVM is a purely visual autoregressive Transformer trained on large-scale image and video data, formulated as sequential “visual sentences.” It has demonstrated the ability to perform diverse visual tasks by adapting to different input patterns through prompting, without task-specific retraining, making it a foundation for our medical video modeling experiments.

\begin{figure*}[t]
	\centering
\includegraphics[width=0.85\linewidth]{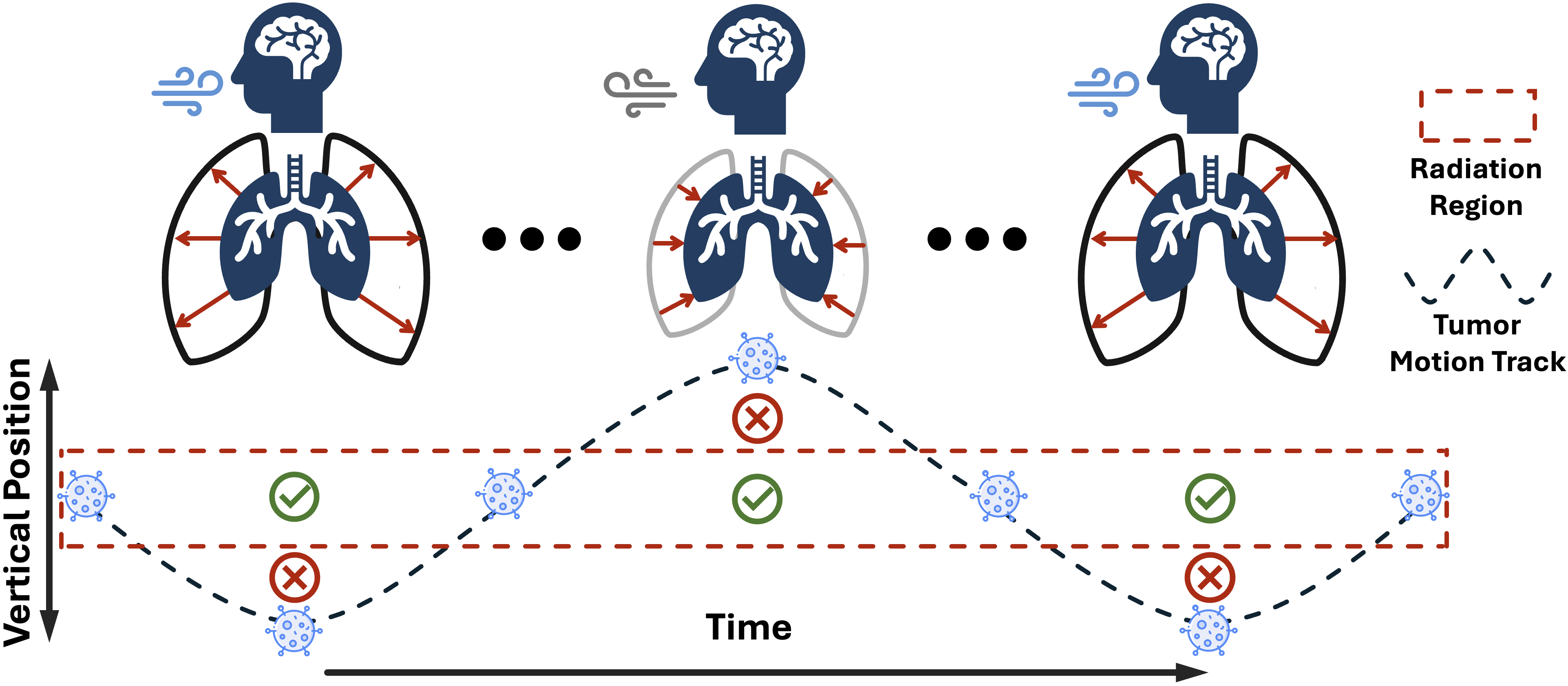}
	\caption{
Schematic illustration of intrafractional tumor motion caused by respiratory cycles during thoracic and upper-abdominal radiotherapy. The top panel depicts the periodic expansion and contraction of the lungs, which drives not only pulmonary tumors but also displaces nearby organs such as the liver and heart, resulting in complex, predominantly vertical motion trajectories (blue dashed curve). The static radiation field (red dashed rectangle) is conventionally planned to encompass the entire tumor motion track to avoid geographic miss, but this inevitably irradiates more surrounding healthy tissue. The green checkmarks (\ding{51}) and red crosses (\ding{55}) mark tumor positions that are either fully covered or missed by the treatment field at different respiratory phases. This figure highlights a key clinical challenge: without accurate motion modeling, margins must be expanded to ensure target coverage for lung, liver, and cardiac-adjacent tumors, which increases unnecessary dose to adjacent organs-at-risk. Precise prediction of tumor trajectories can enable motion-adaptive strategies (e.g., gating, tracking) that safely reduce margins and support patient-specific motion management across multiple thoracic and abdominal sites.
}
    
    \label{fig_main_idea}
\end{figure*}

We evaluate this general-purpose video model, which has never been exposed to medical data, on five representative medical imaging tasks: segmentation, detection, denoising, super-resolution, and motion prediction. Remarkably, despite being trained entirely on natural image and video datasets, the model achieves competitive performance across these diverse tasks. In particular, for motion prediction in radiotherapy, LVM even surpasses specialized baselines by accurately forecasting future 3D CT phases based on prior ones. This improvement highlights the model’s ability to infer patient-specific motion patterns and reason over temporal dynamics, demonstrating strong zero-shot modeling and emergent reasoning capabilities in medical imaging.
We summarize our findings as follows:

\begin{enumerate}
    \item \textbf{Unified video models can address diverse medical imaging tasks.}  
    Our experiments demonstrate that a single large video model can be applied to a variety of medical imaging tasks—including segmentation, detection, denoising, super-resolution, and motion prediction—without task-specific retraining. Although its performance does not yet reach state-of-the-art levels across all tasks, these results highlight the feasibility of developing a unified medical imaging framework through the video modeling pipeline, thereby reducing reliance on highly specialized architectures and handcrafted task designs (\autoref{fig_example_tasks}).
    
    \item \textbf{Video models are particularly effective for sequential prediction tasks.}  
    Even in a zero-shot setting, the large vision model outperforms task-specific baselines such as deformation vector field (DVF)-based and generative models in motion prediction (\autoref{fig_main_idea}, \autoref{tab:performance}). This finding suggests that autoregressive video models inherently capture temporal dependencies and dynamic patterns, making them well-suited for medical applications involving sequential data—such as 4D CT, functional MRI, and dynamic ultrasound—where temporal coherence and motion reasoning are critical.
    
    \item \textbf{Emergent general visual reasoning.}  
    We observe early signs of general visual reasoning, as the model can handle out-of-distribution (OOD) medical data and perform novel tasks without task-specific retraining. Despite never being exposed to medical images during pretraining, it generalizes across modalities and reasoning types, suggesting that large-scale video models can develop transferable representations beyond their training domains. Further investigation is needed to clarify this generalization mechanism and to assess whether pretraining on large-scale medical sequential data could further enhance reasoning and clinical applicability.
\end{enumerate}

\section{Related Works}

\smallskip\noindent\textbf{Video Foundation Models.}  
Recent progress in large-scale video modeling has shown that autoregressive and diffusion-based architectures can learn rich spatiotemporal representations directly from raw video sequences~\cite{yan2021videogpt,ho2022imagen,bai2024sequential}. Inspired by the scaling trends observed in large language models (LLMs) and vision-language models (VLMs), modern video foundation models are trained with simple generative objectives—such as next-frame prediction, masked frame reconstruction, or video continuation—on massive web-scale datasets.  
This paradigm has led to the emergence of \emph{zero-shot} and \emph{few-shot} capabilities, enabling these models to perform diverse visual tasks—including segmentation, tracking, physical reasoning, and scene manipulation—without explicit supervision~\cite{wiedemer2025video}.  
Recent releases such as Veo 3~\cite{google2025veo3announcement,google2025veo3launch} and Sora 2~\cite{openai2025sora2} exemplify this trend, demonstrating that a single generative video model can generalize across perceptual, physical, and reasoning tasks purely through spatiotemporal sequence learning. These results suggest that large-scale video models are on a trajectory toward becoming unified, general-purpose vision foundation models, paralleling how LLMs unified natural language understanding and reasoning.

\smallskip\noindent\textbf{Medical Foundation Models.}  
In the medical domain, foundation models have rapidly evolved from task-specific architectures toward unified multi-modal systems that integrate image, text, and structured data understanding~\cite{sellergren2025medgemma,moor2023med,zhang2023biomedclip,lai2025med}.  
Early medical vision-language models (VLMs) such as BioViL~\cite{boecking2022making} and MedCLIP~\cite{wang2022medclip} learned cross-modal alignment between radiology images and clinical reports, while later systems like Med-Flamingo~\cite{moor2023med} and MedGemma~\cite{sellergren2025medgemma} demonstrated unified reasoning across multiple modalities and tasks, including visual question answering, report generation, and disease classification.  
However, despite these advances, most medical foundation models remain confined to static imaging tasks and lack explicit temporal modeling. Dynamic medical data—such as 4D CT, MRI cine sequences, and ultrasound videos—contain complex motion patterns driven by physiological processes like respiration and cardiac cycles. Exploring whether large-scale video models can generalize zero-shot to these temporally rich modalities is thus an important next step toward true general-purpose medical AI.

\smallskip\noindent\textbf{Medical Motion Modeling.}  
Organ motion modeling plays a central role in radiation therapy and interventional imaging, as it allows for precise tracking and compensation of respiratory-induced deformation.  
Traditional approaches rely on statistical shape and motion models, where principal component analysis (PCA) is applied to deformation vector fields (DVFs) derived from deformable image registration~\cite{nie2013site,vile2015statistical}. While these models provide low-dimensional motion representations, they assume linear motion patterns and are highly sensitive to registration quality.  
Subsequent deep generative methods, including variational autoencoders (VAEs)~\cite{van2017neural} and diffusion models~\cite{rombach2022high,smolders2024diffusert}, introduced non-linear, data-driven motion representations but still depended on precomputed DVFs for supervision, making them vulnerable to error accumulation.  
Sequential prediction models such as ConvLSTM~\cite{shi2015convolutional,ghasemi2024feasibility} and hybrid simulators like RMSim~\cite{lee2023rmsim} capture temporal dynamics more effectively but remain domain-specific and require handcrafted supervision.  

\smallskip\noindent\textbf{Zero-shot Video Models for Medical Motion.}  
Unlike traditional methods, zero-shot video foundation models offer a unified, domain-agnostic framework for spatiotemporal reasoning. Without any retraining, these models can directly process medical 4D CT sequences as temporal input, learning to infer respiratory motion, organ deformation, and temporal coherence from natural video priors.  
This ability to reason over unseen modalities and capture physiological dynamics purely through autoregressive or diffusion-based mechanisms highlights the potential of video models as the next generation of medical foundation models—capable of unifying segmentation, tracking, and motion prediction within a single zero-shot pipeline.

\section{Motion Modeling and Reasoning}\label{sec:autoregressive}

\smallskip\noindent\textbf{Overview.}  
We formulate organ motion prediction as a video modeling and generation problem. In radiotherapy, 4D CT scans capture the dynamics of organ motion by acquiring a series of 3D CT volumes $X_t$ at different time points $t$ within a single respiratory cycle~\cite{hugo2017longitudinal,kwong2015f}. Let $\mathcal{T} = \{0, 1, \dots, T\}$ denote the full set of temporal indices, where $T$ is the total number of breathing phases (typically $T=9$ or $10$). Each volume $X_t$ corresponds to a specific breathing phase and collectively these form temporal sequence $\{X_0, X_1, \dots, X_T\}$ that encodes organ motion.

The primary objective is to learn patient-specific motion patterns from past imaging phases and generate accurate predictions of future phases to support treatment planning and dose delivery. Rather than predicting only the next phase in isolation, the temporal trajectory itself can be modeled as an \emph{autoregressive process} with joint probability:
\begin{align}
P(X_1, X_2, \dots, X_T) = \prod_{t=1}^{T} p_\theta(X_t' \mid X_0, X_1, \dots, X_{t-1})
\end{align}
Here, $X_t'$ denotes the predicted CT volume at phase $t$, and $p_\theta$ is the conditional probability distribution modeled by a neural network with learnable parameters $\theta$. The network can thus make predictions conditioned on all previously observed (or generated) phases, enabling it to capture both short-term and long-range temporal dependencies in respiratory motion.

In practice, suppose we are given a partial sequence of observed phases $\{X_0, X_1, \dots, X_L\}$, where $L<T$. We can generate the remaining future phases $\{X_{L+1}', \dots, X_T'\}$ by maximizing the conditional likelihood:
\begin{align}
\max_{\theta} \prod_{t=L+1}^{T} p_\theta(X_t' \mid X_0, \dots, X_{t-1}')
\end{align}
This formulation aligns naturally with autoregressive generation tasks in natural language processing and time-series prediction, allowing the direct adoption of \emph{autoregressive generation models} that have demonstrated strong performance in learning complex temporal dynamics. Crucially, this approach offers three core advantages:  
(1) It enables explicit modeling of temporal continuity and motion progression across multiple phases;  
(2) It allows each prediction to incorporate information from all previously observed and generated phases, promoting consistency and reducing error accumulation;  
(3) It directly leverages imaging data without relying on intermediate deformation vector fields, allowing the model to learn motion patterns grounded in anatomical appearances and spatial context.
With this assumption, our method is illustrated as follows.

% \begin{figure*}[t]
% 	\centering
% \includegraphics[width=0.9\linewidth]{fig_pipeline.png}
% \caption{\textbf{Video model pipeline.} Each phase of a 4D CT scan is arranged as an ordered input sequence ($X_{0}$ to $X_{T}$), capturing organ motion across the respiratory cycle. A VQGAN-based vision encoder converts each CT image into discrete visual tokens, which are flattened into a 1D sequence to form a compact latent representation. These tokens are fed into an autoregressive transformer that models temporal dependencies across phases and predicts the token set for the next unseen phase conditioned on all preceding ones. This process is repeated recursively to predict the full sequence of future motion states. Finally, the predicted tokens are decoded by the VQGAN decoder to reconstruct CT images ($X'_{0}$ to $X'_{T}$) that capture the anticipated anatomy at each future phase. By combining discrete tokenization, temporal autoregression, and iterative prediction, LVM enables high-fidelity, patient-specific modeling of respiratory-induced motion patterns in 4D CT.}

%     \label{fig_pipeline}
% \end{figure*}

\subsection{Pipeline}

\smallskip\noindent\textbf{Pre-processing.}  
To enable structured modeling of respiratory motion, each 3D volume in the 4D CT sequence is first segmented to isolate clinically relevant structures. We employ nnUNet~\cite{isensee2021nnu}, a state-of-the-art self-configuring segmentation framework, initialized with weights from TotalSegmentator~\cite{wasserthal2023totalsegmentator}. This setup allows robust automatic segmentation of lungs, heart, and liver across varying patient anatomies and imaging conditions. nnUNet dynamically adjusts its architecture, input patch size, and training protocols based on data statistics, achieving reliable performance without manual tuning.

The resulting voxel-wise organ masks serve two essential purposes: (i) they provide geometric priors that guide motion modeling in later stages; and (ii) they serve as ground truth for evaluating prediction accuracy via segmentation-based metrics. By comparing the predicted and reference masks across phases, we can quantitatively assess how well the model preserves organ shape, location, and boundary continuity throughout the breathing cycle.

\smallskip\noindent\textbf{CT tokenization.}
To adapt transformer-based autoregressive modeling to CT images, prior vision-language models have often divided images into fixed patches and flattened them into 1D token sequences~\cite{dosovitskiy2020image}. Instead of patch embedding, LVM adopts a discrete latent representation via vector quantization (VQ)~\cite{van2017neural,lai2023memory}, which encodes each CT image into a compact sequence of learnable tokens while preserving spatial structure. 

As illustrated by \citet{bai2024sequential}, LVM integrates a vector-quantized generative adversarial network (VQGAN)~\cite{esser2021taming} into the pipeline. VQGAN consists of a CNN-based encoder, decoder, and a discrete codebook. The encoder compresses input CT slices into a latent feature map through multiple downsampling stages; this latent grid is quantized into discrete codebook entries, yielding a $16\times16$ grid of tokens (256 tokens for each $256\times256$ CT slice). The decoder then reconstructs CT images from predicted token sequences through aligned upsampling layers. This tokenization strategy significantly reduces sequence length compared to raw pixel modeling, making autoregressive training tractable while retaining anatomical representations.

\smallskip\noindent\textbf{Sequence modeling of CT phases.}  
Once tokenized, each CT phase in the 4D sequence becomes a temporally ordered grid of discrete visual tokens. LVM models this sequence using a unidirectional decoder-only transformer with causal self-attention~\cite{vaswani2017attention}. Inspired by autoregressive language models~\cite{brown2020language}, the transformer generates the token sequence for each future CT phase one step at a time, conditioned on all previously seen (or predicted) tokens.

During training, the model is exposed to the complete sequence of phases $\{X_0, ..., X_T\}$ and learns to predict the token grid of phase $X_t$ from preceding phases $\{X_0, ..., X_{t-1}\}$. During inference, only an initial prefix ($\{X_0, ..., X_L\}$) is observed, and the model autoregressively predicts future motion phases. By feeding prior CT phases as sequential context, the transformer progressively predicts token sequences for subsequent phases, enabling multi-phase motion prediction rather than single-step prediction. This formulation allows the model to learn patient-specific respiratory patterns directly from image sequences, leading to more temporally consistent and anatomically plausible phase generation.

\subsection{Implementation Details}  

Organ masks for the lungs, heart, and liver were extracted using the nnUNet model provided by the TotalSegmentator toolkit~\cite{wasserthal2023totalsegmentator}. nnUNet is a UNet-based segmentation framework~\cite{ronneberger2015u} that automatically configures its architecture, patch size, and training parameters based on dataset-specific characteristics~\cite{li2024abdomenatlas,lai2024pixel,chen2024analyzing}. The resulting segmentation masks were used for evaluation and as auxiliary labels to assess motion prediction accuracy.

For image tokenization, we employed a vector-quantized generative adversarial network (VQGAN)~\cite{esser2021taming} with a downsampling factor of \(f=16\) and a codebook size of 8192. Each \(256 \times 256\) CT slice is compressed into a \(16 \times 16\) latent grid (256 tokens). These tokens are then flattened into a 1D sequence, producing a compact and discrete representation for each CT phase. Pre-trained VQGAN parameters from Yutong \etal~\cite{bai2024sequential} were adopted and fine-tuned on our 4D CT data to ensure robust encoding and reconstruction quality.

LVM uses an autoregressive modeling framework built on a decoder-only Transformer based on the LLaMA architecture~\cite{touvron2023llama}. To ensure reproducibility, we explicitly detail the configuration: the model comprises 12 Transformer layers, each with 16 self-attention heads and an embedding dimension of 1024. The feed-forward layers use an expansion factor of 4 (MLP hidden dimension of 4096). Rotary positional embeddings (RoPE) are applied to preserve the temporal order of tokens across phases, and causal attention masking ensures that each token is predicted only from preceding tokens. In total, the model contains approximately 120 million parameters and supports a context length of 4096 tokens, corresponding to up to 16 CT phases ($16\times256$ tokens) within the VQGAN tokenizer framework. This architecture enables the model to capture extended intra-fraction temporal dependencies while remaining computationally tractable.

% \subsection{Implementation Details}

% Organ masks for the lungs and heart are extracted using the nnUNet model from the TotalSegmentator toolkit~\cite{wasserthal2023totalsegmentator}, which is based on the UNet architecture~\cite{ronneberger2015u} and automatically adjusts hyperparameters according to dataset characteristics~\cite{li2024abdomenatlas,lai2024pixel,chen2024analyzing}. CT images are resampled to an in-plane resolution of \(256 \times 256\) and intensity-normalized to the standard Hounsfield unit (HU) range of \([-1000, 400]\).

% The VQGAN module~\cite{esser2021taming} employs a downsampling factor of \(f=16\) and a codebook size of 8192, producing a \(16 \times 16 = 256\) token grid for each image. Token sequences are flattened in row-major order to form a 1D sequence per CT phase. We use pre-trained parameters from Yutong~\etal~\cite{bai2024sequential} for the VQGAN encoder and decoder.

% The autoregressive transformer is implemented based on the LLaMA architecture~\cite{touvron2023llama}, supporting a context length of up to 4096 tokens, which allows the model to process sequences of up to 16 CT phases in a single pass. The model is optimized using AdamW with an initial learning rate of \(1 \times 10^{-4}\) and a cosine learning rate scheduler. A cross-entropy loss is used to supervise token prediction. Training is performed on four NVIDIA A100 GPUs with a batch size of 4 patients per iteration. Each experiment is trained for 200 epochs, totaling approximately 72 GPU hours for convergence.

\begin{table*}[t]
    \centering
    % \scriptsize
    \caption{\textbf{Performance on Next-Phase Motion Prediction.} LVM significantly outperforms previous methods and achieves high accuracy. We evaluate performance by comparing the segmentation masks of organs between the predicted and ground-truth CT phases. Higher IoU and DSC values indicate more precise overall predictions, while lower SD and HD values suggest more accurate boundary alignment.}

\begin{tabular}{c|cc|cccc}
    \toprule
    Data & Organ & Method & IoU (\%) $\uparrow$ & DSC (\%) $\uparrow$ & SD (px) $\downarrow$ & HD95 (px) $\downarrow$ \\
    \midrule 
   \multirow{10}{*}{Public} & \multirow{5}{*}{Lung} & DAM~\cite{pastor2023probabilistic} & 80.08 & 85.47 & 4.84 & 7.14\\
   & & DiffuseRT~\cite{smolders2024diffusert} & 81.23 & 86.87 & 4.19 & 6.92\\
   & & ConvLSTM~\cite{ghasemi2024feasibility} & 82.36 & 88.14 & 3.92 & 5.76\\
   & & RMSim~\cite{lee2023rmsim} & 84.05 & 89.16 & 3.66 & 5.08\\
   & & Ours& \cellcolor{ired!20} 90.75 & \cellcolor{ired!20}95.15 & \cellcolor{ired!20}2.65 & \cellcolor{ired!20}4.33\\

    \cmidrule{2-7}
    
    & \multirow{5}{*}{Heart} & DAM~\cite{pastor2023probabilistic} & 78.62 & 82.69 & 5.59 & 9.83\\
   & & DiffuseRT~\cite{smolders2024diffusert} & 80.29 & 85.08 & 4.95 & 9.12\\
   & & ConvLSTM~\cite{ghasemi2024feasibility} & 81.41 & 86.27 & 4.41 & 8.45\\
   & & RMSim~\cite{lee2023rmsim} & 82.97 & 87.96 & 3.94 & 7.14\\
   & & Ours& \cellcolor{ired!20}88.43& \cellcolor{ired!20}93.85 & \cellcolor{ired!20} 2.39& \cellcolor{ired!20}4.24\\
    
    \midrule 
   \multirow{15}{*}{Private} & \multirow{5}{*}{Lung} & DAM~\cite{pastor2023probabilistic} & 79.71 & 83.15 & 5.98 & 8.16\\
   & & DiffuseRT~\cite{smolders2024diffusert} & 82.44 & 85.36 & 5.44 & 7.23\\
   & & ConvLSTM~\cite{ghasemi2024feasibility} & 84.06 & 88.44 & 4.54 & 6.81\\
   & & RMSim~\cite{lee2023rmsim} & 85.35 & 89.62 & 3.85 & 6.54\\
   & & Ours&  \cellcolor{ired!20}91.88 & \cellcolor{ired!20}95.74 & \cellcolor{ired!20}2.84 & \cellcolor{ired!20}4.47\\

    \cmidrule{2-7}
    
    & \multirow{5}{*}{Heart} & DAM~\cite{pastor2023probabilistic} & 76.81 & 80.97 & 6.63 & 9.05\\
   & & DiffuseRT~\cite{smolders2024diffusert} & 79.07 & 83.43 & 6.29 & 8.58\\
   & & ConvLSTM~\cite{ghasemi2024feasibility} & 80.54 & 85.25 & 5.77 & 7.59\\
   & & RMSim~\cite{lee2023rmsim} & 82.14 & 87.03 & 5.32 & 7.15\\
   & & Ours& \cellcolor{ired!20}87.83 & \cellcolor{ired!20}93.24 & \cellcolor{ired!20}3.34 & \cellcolor{ired!20}5.38\\

    \cmidrule{2-7}
    
   & \multirow{5}{*}{Liver} & DAM~\cite{pastor2023probabilistic} & 77.15 & 81.46 & 4.99 & 9.34\\
   & & DiffuseRT~\cite{smolders2024diffusert} & 79.18 & 83.04 & 4.77 & 8.63\\
   & & ConvLSTM~\cite{ghasemi2024feasibility} & 82.68 & 86.18 & 4.35 & 8.26\\
   & & RMSim~\cite{lee2023rmsim} & 83.05 & 88.08 & 3.70 & 6.75\\
   & & Ours& \cellcolor{ired!20}91.99 & \cellcolor{ired!20}95.83 & \cellcolor{ired!20} 2.95 & \cellcolor{ired!20}5.09\\
    
    \bottomrule
\end{tabular}
\begin{tablenotes}
    \item Public Lung \& Heart: 80 4D CT scans, 800 3D CT scans; Private Lung \& Heart: 50 4D CT scans, 500 3D CT scans; Private Liver: 52 4D CT scans, 520 3D CT scans.
    \item IoU - intersection over union; DSC - dice similarity coefficient; SD - surface distance; HD - Hausdorff distance.
\end{tablenotes}
\label{tab:performance}
\end{table*}

\section{Experiment}

\subsection{Experiment Setting}

\smallskip\noindent\textbf{Dataset.}  
Our study utilizes two 4D CT datasets. The first dataset, provided by Hugo~\etal~\cite{hugo2017longitudinal}, is publicly available and includes 4D CT scans from 20 patients diagnosed with locally advanced non-small cell lung cancer. All scans were acquired using a 16-slice helical CT scanner (Brilliance Big Bore, Philips Medical Systems, Andover, MA) with respiration-correlated CT imaging. Each scan includes 10 3D CT phases (0\% to 90\%) generated by phase-based binning, with a slice thickness of 3 mm.

The second dataset was collected at the Department of Radiation Oncology, Emory University. Similar to the public dataset, each 4D CT scan consists of 10 breathing phases covering a complete respiratory cycle. Scans were performed using a Siemens SOMATOM Definition AS scanner at 120 kV, following the standard Siemens Lung 4D CT protocol. The imaging system was configured with Syngo CT VA48A software, a pitch of 0.8, and a Bf37 reconstruction kernel. The image resolution is \(512 \times 512 \times (133\text{-}168)\) voxels, with voxel spacing of \(0.9756 \times 0.9756 \times 2.0\) mm\(^3\) along the \(x, y,\) and \(z\) axes.

The third dataset is an in-house liver cancer therapy cohort comprising 52 patients with hepatocellular carcinoma (HCC) who underwent pre-treatment 4D CT scans. All scans were acquired using a [scanner name, e.g., Siemens SOMATOM Force] at 120 kV, following the standard liver 4D CT protocol with respiratory phase-based binning (10 phases). The slice thickness was 2 mm, and the in-plane resolution is \(512 \times 512\) with voxel spacing of [e.g., \(0.976 \times 0.976 \times 2.0\) mm\(^3\)]. Patients were scanned in the supine position under free breathing. Institutional ethics approval was obtained for retrospective data use.

\smallskip\noindent\textbf{Evaluation.}
Our primary goal is to evaluate how well the zero-shot video model can perform medical imaging tasks without domain-specific training. For radiotherapy motion prediction, we focus on assessing the accuracy of predicted organ positions and shapes rather than low-level pixel similarity. Spatial agreement between predicted and ground-truth segmentations is evaluated using four widely adopted metrics: Intersection over Union (IoU), Dice Similarity Coefficient (DSC), Surface Distance (SD), and Hausdorff Distance (HD). IoU and DSC measure volumetric overlap between predicted and reference masks, with DSC being particularly informative for small or irregular structures. SD quantifies the average symmetric distance between organ surfaces, while HD represents the maximum point-to-point deviation, reflecting worst-case boundary errors. Together, these metrics provide a comprehensive view of spatial accuracy and indicate whether radiation can be delivered precisely to the target while sparing nearby healthy tissues. 
For each generated CT phase, organ masks are obtained using the nnUNet~\cite{isensee2021nnu} model, ensuring consistent segmentation quality across all phases and preventing potential bias from using different tools for ground-truth and evaluation.

Beyond motion prediction, we further assess the zero-shot generalization ability of the video model across three representative medical imaging tasks: \textit{segmentation}, \textit{denoising}, and \textit{super-resolution}. 
For segmentation, the model receives CT slices and text prompts specifying target organs (e.g., “segment the liver”) and outputs binary masks, evaluated using Dice and IoU. 
For denoising, synthetically corrupted CT images are restored to clean versions, with performance measured by PSNR and SSIM. 
For super-resolution, low-resolution CT inputs are upscaled to high-resolution predictions, again evaluated by PSNR and SSIM to assess structural fidelity. 
Although the model is not trained on medical data, it achieves stable results across all three tasks, demonstrating its ability to understand anatomical structures, preserve texture continuity, and generalize across distinct medical imaging objectives.

\begin{table}[t]
\centering
\caption{\textbf{Zero-shot performance of the large video model (LVM) across representative medical imaging tasks.}
Each task is evaluated using its most widely adopted quantitative metrics, reflecting the model’s cross-task generalization ability and its potential to inspire the development of the unified medical vision model.}
\label{tab:toy_study}
\begin{tabular}{l|c|c}
\toprule
\textbf{Task} & \textbf{Metric} & \textbf{Zero-shot LVM}  \\
\midrule
\multirow{2}{*}{Segmentation} & DSC & 91.52 \\
& IoU & 87.12\\
\midrule
\multirow{2}{*}{Super-resolution} & SSIM & 78.35 \\
& PSNR & 29.56 \\
\midrule
\multirow{2}{*}{Denoising} & SSIM  & 85.38 \\
& PSNR & 34.44 \\
\bottomrule
\end{tabular}
\end{table}

\subsection{Results \& Analysis}

\begin{figure*}[t]
	\centering
\includegraphics[width=0.8\linewidth]{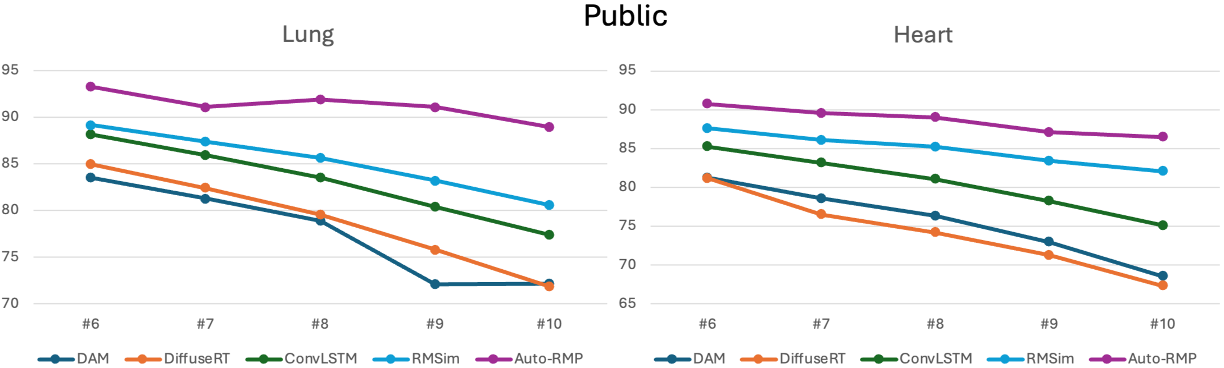}
\caption{\textbf{Multi-phase motion prediction on the public dataset.} 
We evaluate model performance on the public 4D CT dataset using Dice Similarity Coefficient (DSC, \%). Each model is provided with the first five phases of the 4D CT scan and autoregressively predicts the next five phases. The plots show phase-by-phase DSC for five representative methods (DAM, DiffuseRT, ConvLSTM, RMSim, and our proposed LVM). LVM consistently achieves the highest DSC across all predicted phases and exhibits the smallest performance drop from phase \#1 to \#5, indicating its superior ability to model smooth and realistic multi-phase motion patterns.}

    \label{fig_longterm_public}
\end{figure*}

\begin{figure*}[t]
	\centering
\includegraphics[width=\linewidth]{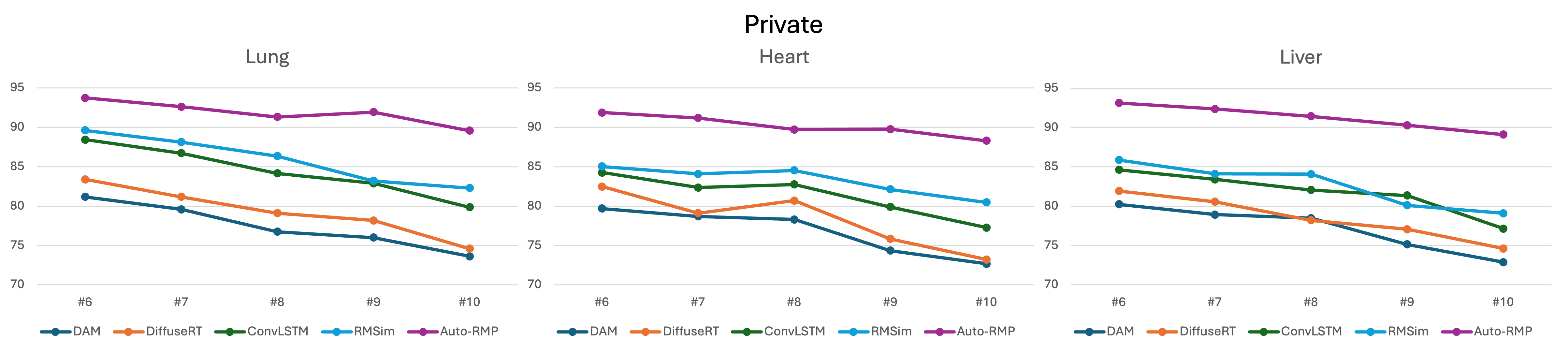}
\caption{\textbf{Multi-phase motion prediction on the private dataset.} 
The same DSC-based evaluation is conducted on our institutional 4D CT dataset (including lung, heart, and liver cases). Each model receives the first five phases and must generate the subsequent five phases. LVM maintains consistently higher DSC across all organs and phases, with smoother phase-to-phase transitions and less degradation compared to competing methods, demonstrating strong robustness and generalization to in-house data.}

    \label{fig_longterm_private}
\end{figure*}

\subsubsection{Segmentation, Denoising, and Super-Resolution}

We further evaluate the zero-shot capability of the large video model (LVM) on three representative medical imaging tasks: organ segmentation, image denoising, and super-resolution. These tasks collectively cover structural understanding, low-level enhancement, and spatial fidelity—key components of clinical image interpretation.

As shown in \tableautorefname~\ref{tab:toy_study}, the zero-shot LVM achieves promising performance across all tasks without any domain-specific training. For segmentation, the model produces accurate organ boundaries with a Dice score of 91.52\%, demonstrating its ability to localize anatomical structures directly from CT inputs. For denoising and super-resolution, LVM attains PSNR values of 34.4~dB and 29.6~dB, respectively, with SSIM around 0.8 in both cases, indicating perceptual quality and structure preservation. 

While these results remain below the best supervised or fine-tuned models, they are notable given the absence of medical data during training. The LVM demonstrates strong generalization and the ability to infer spatial priors, texture consistency, and anatomical structure in a zero-shot setting. These findings highlight the potential of video models as general-purpose medical imaging backbones—capable of transferring across tasks and modalities.

\subsubsection{Next-Phase Motion Prediction.}
We evaluated \textit{LVM} against recent deep learning methods for organ motion prediction, including DAM~\cite{pastor2023probabilistic}, DiffuseRT~\cite{smolders2024diffusert}, ConvLSTM~\cite{ghasemi2024feasibility}, and RMSim~\cite{lee2023rmsim}. As shown in \tableautorefname~\ref{tab:performance}, \textit{LVM} consistently outperformed all baselines across both public and private datasets.

On the public dataset, \textit{LVM} achieved an IoU of 90.75\% and a DSC of 95.15\% for lung motion prediction, exceeding the best baseline (RMSim) by 5.6\% (IoU) and 5.1\% (DSC). For heart motion, it reached an IoU of 88.43\% and a DSC of 93.85\%, and further improved boundary alignment, with lower SD and HD compared to RMSim .

On the private dataset, which includes more challenging and heterogeneous motion patterns, \textit{LVM} maintained strong performance. It achieved an IoU of 91.88\% and a DSC of 95.74\% for lung motion prediction, outperforming the best baseline by 6.3\% (IoU) and 5.9\% (DSC). For heart motion, \textit{LVM} achieved an IoU of 87.83\% and a DSC of 93.24\%, with clear reductions in surface distance and Hausdorff distance. For liver motion—a particularly difficult organ due to its irregular deformation—\textit{LVM} achieved an IoU of 91.99\% and a DSC of 95.83\%, surpassing the next-best model by 8.7\% (IoU) and 7.0\% (DSC).
These results highlight two findings:
(1) \textit{LVM} produces segmentation masks with higher spatial agreement (IoU/DSC) and more accurate boundary alignment (lower SD/HD) than previous methods; and
(2) These gains are consistent across organs and datasets, demonstrating generalizability under diverse imaging conditions.

We attribute this performance advantage to the autoregressive design of \textit{LVM}. Existing generative approaches (e.g., DAM~\cite{pastor2023probabilistic}, DiffuseRT~\cite{smolders2024diffusert}) often rely on limited conditioning information (typically a single CT phase), which restricts their ability to represent the full temporal dynamics of breathing. Sequence-based models like ConvLSTM~\cite{ghasemi2024feasibility} partially address this by propagating temporal states, while physics-based simulators like RMSim~\cite{lee2023rmsim} embed predefined assumptions about motion fields. However, these methods either lack strong mechanisms for long-sequence consistency or are constrained by their assumptions.

\begin{table*}[t]
    \centering
    % \scriptsize
    \caption{\textbf{Ablation study on input modalities.} 
We evaluate the impact of different input modalities on LVM performance using three configurations: (i) \emph{CT Only}—raw CT images without any segmentation guidance, (ii) \emph{Mask Only}—organ segmentation masks providing anatomical structure but no texture information, and (iii) \emph{CT + Mask}—a combined input that integrates both spatial texture and anatomical priors. Performance is assessed on lung and heart using a mixed dataset (20 public + 20 private 4D CT scans) and reported across four metrics: IoU, DSC, SD, and HD. The combination of CT and masks consistently yields the best performance across all organs and metrics, demonstrating that anatomical structure and image appearance provide complementary cues that jointly enhance motion modeling accuracy and boundary fidelity.}

\begin{tabular}{c|cc|cccc}
    % {ll|cc|cc|cc}
        \toprule
        Data & Organ & Input & IoU (\%) $\uparrow$ & DSC (\%) $\uparrow$ & SD (px) $\downarrow$ & HD (px) $\downarrow$ \\
        \midrule 
       \multirow{6}{*}{Mixed} & \multirow{3}{*}{Lung} & CT Only & 85.83 & 88.57 & 3.98 & 4.86\\
       & & Mask Only & 87.60 & 91.88 & 3.19 & 4.29\\
       & &  CT + Mask & \cellcolor{ired!20}91.23 & \cellcolor{ired!20}94.87 & \cellcolor{ired!20}2.75 & \cellcolor{ired!20}4.18\\

        \cmidrule{2-7}
        
        & \multirow{3}{*}{Heart} & CT Only & 83.71 & 86.83 & 4.11 & 5.93\\
       & & Mask Only & 84.25 & 88.12 & 3.29 & 5.58\\
       & & CT + Mask & \cellcolor{ired!20}87.38 & \cellcolor{ired!20}91.11 & \cellcolor{ired!20}2.87 & \cellcolor{ired!20}4.81\\

       \cmidrule{1-7}

       \multirow{3}{*}{Private} & \multirow{3}{*}{Liver}  & CT Only & 86.46 & 92.74 & 3.66 & 4.40 \\
       & & Mask Only & 88.64 & 93.97 & 3.22 & 4.12 \\
       & & CT + Mask & \cellcolor{ired!20}92.95 & \cellcolor{ired!20}96.35 & \cellcolor{ired!20}2.95 & \cellcolor{ired!20}3.98 \\
       
        \bottomrule
    \end{tabular}

    \begin{tablenotes}
        \item Mixed: 20 4D CT scans of public dataset, 20 4D CT scans of our private dataset
        \item IoU - intersection over union; DSC - dice similarity coefficient; SD - surface distance; HD - Hausdorff distance.
    \end{tablenotes}

    \label{tab:ablation}
\end{table*}

In contrast, \textit{LVM} conditions each prediction on a sequence of past CT phases and feeds its predictions into subsequent steps, creating a feedback loop. This design allows the model to catch patient-specific motion patterns and maintain temporal consistency across predicted phases, which closely aligns with the requirements of 4D CT motion prediction.

\subsubsection{Multi-phase Motion Prediction.}
The multi-phase motion prediction task provides a rigorous benchmark for evaluating a video model’s ability to reason about temporal dynamics and maintain long-term consistency in a purely zero-shot setting. As shown in Figures~\ref{fig_longterm_public} and~\ref{fig_longterm_private}, baseline models exhibit a steady decline in Dice Similarity Coefficient (DSC) as predictions extend across future CT phases—highlighting the well-known challenge of error accumulation in sequential motion forecasting.

Among supervised baselines, ConvLSTM and RMSim perform better than diffusion- or GAN-based models (e.g., DiffuseRT, DAM) because they incorporate either sequential recurrence or physics-informed priors. However, these approaches depend on deformation vector field (DVF) supervision and never directly observe CT appearance, limiting their ability to learn patient-specific respiratory dynamics or spatially coherent motion patterns.

In contrast, the zero-shot LVM achieves strong temporal coherence and spatial fidelity across all datasets, maintaining DSC above 85\% even at the fifth predicted phase. Despite never being trained on medical data, the model learns to infer and reason about organ motion trajectories directly from image sequences—capturing both periodic breathing patterns and smooth anatomical deformation. Its autoregressive design allows previously predicted frames to serve as temporal context, effectively reducing drift and maintaining realistic continuity over long horizons.

These findings reveal that large video models, even in zero-shot settings, can exhibit emergent abilities in medical reasoning—integrating visual understanding, memory, and prediction across multiple phases. This suggests that video models hold strong potential as unified, general-purpose backbones for medical video analysis, capable of supporting downstream applications like motion compensation, adaptive radiotherapy planning, and 4D image synthesis.

% \subsubsection{Ablation Study.}  
% To evaluate the contribution of different input modalities, we conduct an ablation study comparing three configurations: (i) CT images only, (ii) segmentation masks only, and (iii) a combination of both (CT + Mask). As summarized in \tableautorefname~\ref{tab:ablation}, using only raw CT inputs captures global anatomy but lacks strong localization cues for specific organ boundaries, resulting in lower IoU and DSC scores. In contrast, using only segmentation masks enables better boundary localization and more focused motion modeling, but omits the rich spatial and intensity context from the CT image.

% Combining both modalities yields consistently superior results across all metrics and organs. The CT + Mask configuration achieves the highest DSC and IoU, while also reducing boundary-related errors as indicated by lower SD and HD. This improvement demonstrates that segmentation masks provide anatomical priors that guide the network to focus on clinically relevant regions, while CT intensities provide fine-grained spatial context to refine motion details. These results validate our design choice of integrating CT images with segmentation masks in LVM, allowing the model to simultaneously learn appearance features and structural priors for more accurate and robust motion prediction.

\begin{figure*}[t]
	\centering
\includegraphics[width=\linewidth]{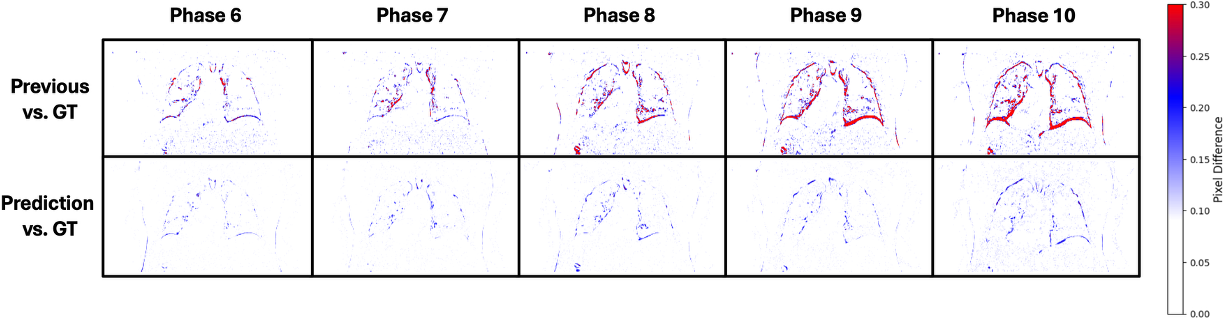}
\vspace{-20pt}
\caption{
\textbf{Qualitative visualization of lung motion.} The first five phases are used as input, and the model predicts the next five. Each heatmap shows voxel-wise pixel differences between the ground truth (GT) and either the previous phase or the model prediction. Red indicates larger discrepancies. LVM accurately captures respiratory-induced motion, showing reduced errors and smoother temporal transitions compared to the prior phase, demonstrating coherent and anatomically consistent lung motion prediction.}
    \label{fig_vis_emory_lung}
\end{figure*}

% \begin{figure*}[t]
% 	\centering
% \includegraphics[width=\linewidth]{fig_vis_emory_lung50.png}
% \caption{
% \textbf{Qualitative visualization of lung motion.} }
%     \label{fig_vis_emory_lung50}
% \end{figure*}

% \begin{figure*}[t]
% 	\centering
% \includegraphics[width=\linewidth]{fig_vis_nih_lung.png}
% \caption{
% \textbf{Qualitative visualization of lung motion.} }
%     \label{fig_vis_nih_lung}
% \end{figure*}

\begin{figure*}[t]
	\centering
\includegraphics[width=0.8\linewidth]{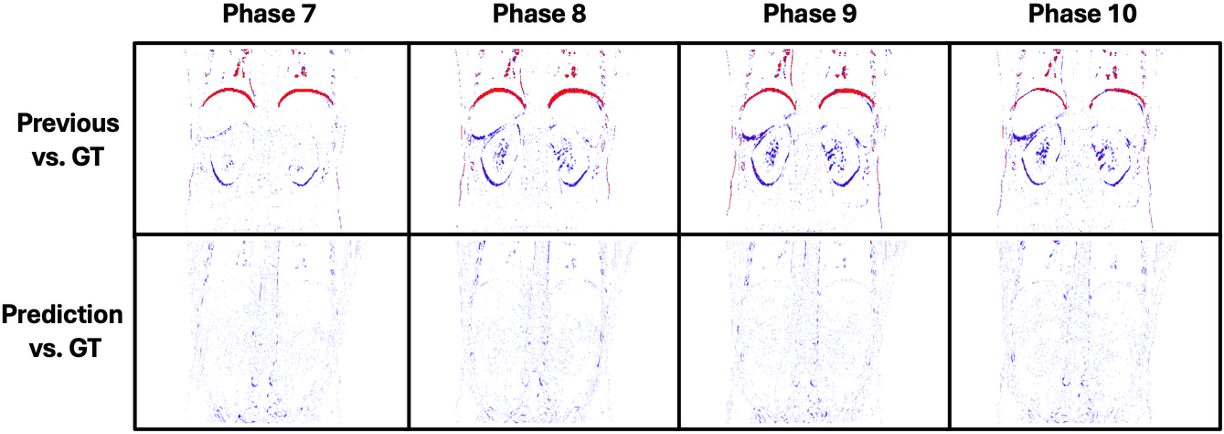}
\caption{
\textbf{Qualitative visualization of liver motion.} The first five 4D CT phases are used as input, and the model predicts the next five. Each heatmap shows voxel-wise differences between the prediction (or previous phase) and the ground truth, where red indicates larger errors. LVM accurately captures the liver’s smooth deformation and diaphragm-induced motion, maintaining temporal and anatomical consistency across phases.}
    \label{fig_visulization_liver}
\end{figure*}

\subsubsection{Ablation Study.}
To better understand how different visual cues influence the zero-shot reasoning ability of the video model, we conduct an ablation study comparing three input settings: (i) \textit{CT Only}, where the model learns directly from raw CT intensities; (ii) \textit{Mask Only}, where the input is limited to segmentation masks highlighting organ structures; and (iii) \textit{CT + Mask}, our default configuration combining both modalities. Results are summarized in \tableautorefname~\ref{tab:ablation} across lung, heart, and liver datasets.

With \textit{CT Only}, the model leverages texture and intensity information to capture local deformation patterns, but lacks explicit structural guidance. This often results in blurred boundaries or overestimation of motion in background regions (e.g., 88.57\% DSC for lung).
When using \textit{Mask Only}, the model focuses on organ contours and motion-specific regions, improving boundary alignment but losing fine-grained appearance cues essential for accurate deformation estimation.
Combining both (\textit{CT + Mask}) yields the best performance across all metrics (e.g., 94.87\% DSC for lung, 96.35\% for liver), demonstrating that CT intensities and segmentation masks provide complementary information: CT images encode texture and density continuity, while masks constrain motion learning to anatomically relevant regions.

These findings suggest that even in a zero-shot setting, large video models can effectively integrate multi-modal spatial cues to reason about organ structure and dynamics. The results highlight that combining pixel-level and semantic-level representations enhances both anatomical fidelity and temporal coherence, offering a more robust foundation for medical video understanding.

% \begin{figure*}[t]
% 	\centering
% \includegraphics[width=\linewidth]{fig_visualization.png}
% \caption{
% \textbf{Qualitative visualization of lung motion.} Coronal views of lung 4D CT are shown, where the top row represents input CT fractions and the bottom row shows predicted future fractions by LVM. The red dashed line serves as a visual reference to highlight motion caused by respiratory cycles. The predicted frames closely follow the patient-specific breathing pattern, demonstrating LVM’s ability to generate temporally consistent and anatomically plausible motion trajectories.
% }
%     \label{fig_visulization_lung}
% \end{figure*}

% \begin{figure*}[t]
% 	\centering
% \includegraphics[width=\linewidth]{fig_visualization_liver.png}
% \caption{
% \textbf{Qualitative visualization of liver motion.} 
% The top row shows the input liver 4D CT fractions, the middle row presents the predicted future CT fractions by LVM, and the bottom row overlays the predicted CT images with ground-truth liver segmentations masks. The close alignment between predicted structures and reference masks demonstrates LVM’s ability to model organ-specific motion accurately over time.
% }
%     \label{fig_visulization_liver}
% \end{figure*}

\begin{figure*}[t]
	\centering
\includegraphics[width=0.9\linewidth]{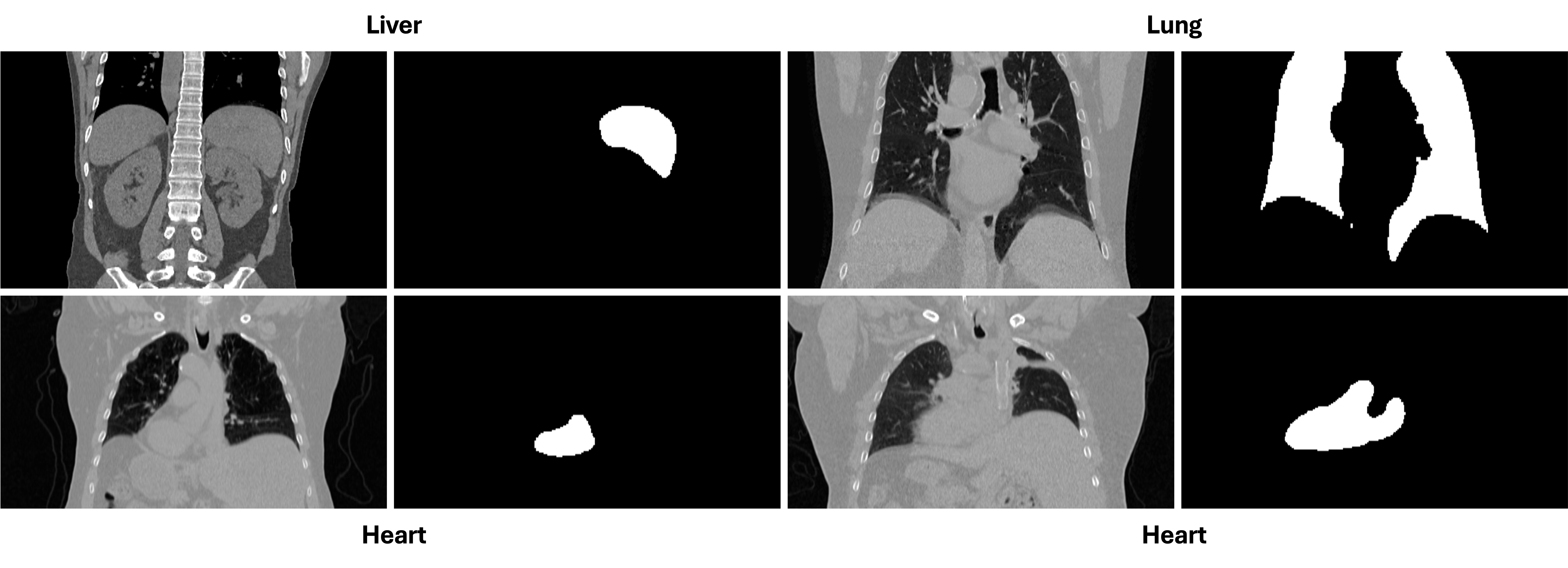}
\caption{
\textbf{Qualitative visualization of Segmentation.} For each organ, the left column shows the original CT slice, and the right column shows the predicted segmentation mask. The results demonstrate that the zero-shot video model can accurately segment organs across diverse anatomical regions based on the given input prompts.}
    \label{fig_seg}
\end{figure*}

\subsubsection{Visualization}

To qualitatively evaluate the effectiveness of LVM in modeling organ motion, we present representative visualizations of predicted versus ground-truth CT phases for different anatomical regions. While quantitative metrics such as DSC and HD provide overall accuracy, qualitative visualization enables intuitive assessment of temporal consistency, local deformation patterns, and anatomical boundary fidelity—factors critical for clinical interpretability and radiotherapy planning. We highlight examples for lung and liver 4D CT sequences, demonstrating LVM's capacity to generate smooth, realistic motion pattern aligned with patient-specific breathing dynamics.

\smallskip\noindent\textbf{Lung 4D CT Motion Prediction.}
\autoref{fig_vis_emory_lung} presents a qualitative heatmap analysis of lung motion prediction for a representative thoracic radiotherapy case. The first five 4D CT phases are used as input, and the model predicts the next five phases. Each heatmap shows voxel-wise pixel differences between the predicted or previous phase and the ground truth (GT), with red indicating larger discrepancies. Compared to the prior phase, the prediction from LVM demonstrates significantly reduced errors across lung boundaries and internal structures, indicating accurate modeling of respiratory-induced expansion and contraction. The results highlight LVM’s ability to learn patient-specific breathing dynamics and maintain temporal and anatomical consistency across multiple phases—key requirements for motion-adaptive radiotherapy planning.

\smallskip\noindent\textbf{Liver 4D CT Motion Prediction.}
\autoref{fig_visulization_liver} shows a qualitative heatmap analysis of liver motion prediction under respiratory influence. Similar to the lung case, the model is provided with the first five 4D CT phases and tasked to predict the next five. Each heatmap represents voxel-wise differences between the predicted or previous phase and the ground truth (GT), where red indicates higher pixel deviation. LVM effectively captures the smooth deformation and positional shift of the liver caused by diaphragm motion, showing substantially reduced discrepancies compared to the prior phase. This demonstrates the model’s capability to reason about complex abdominal motion while maintaining anatomical fidelity over time critical requirement for accurate dose planning and delivery in liver radiotherapy.

\smallskip\noindent\textbf{Segmentation.}
\autoref{fig_seg} shows qualitative examples of zero-shot organ segmentation for the liver, heart, and lung. Without any task-specific fine-tuning, the video model accurately identifies anatomical structures and delineates organ boundaries directly from CT images. These results suggest that the model can generalize to spatial reasoning tasks in medical imaging, effectively transferring its visual understanding from natural scenes to clinical contexts.

These qualitative visualizations highlight the potential of the model to advance motion management by providing predictions with patient-specific motion patterns, which are critical for adaptive treatment planning.

\section{Limitations \& Future Directions}

\noindent
Although our study provides the first evidence that zero-shot video models can perform clinically meaningful reasoning on medical imaging data, several challenges remain before such models can be reliably integrated into clinical radiotherapy workflows.

\smallskip
\noindent\textbf{Beyond intra-fraction motion.}
Our current work primarily focuses on intra-fraction motion prediction, capturing respiratory-induced organ motion within a single 4D CT session. However, real-world radiotherapy treatments span multiple weeks, during which inter-fraction anatomical changes such as tumor shrinkage, weight loss, and baseline organ shifts frequently occur. Modeling these long-term variations will require incorporating longitudinal imaging across multiple sessions and reasoning over extended temporal contexts. Extending video models to handle such inter-fraction dynamics represents an important direction for future research.

\smallskip
\noindent\textbf{Generalization across fundamental imaging tasks.}
Our exploration of segmentation, denoising, and super-resolution tasks demonstrates that large video models can already generalize across fundamentally different medical imaging problems in a zero-shot setting. Although the current performance does not yet match specialized, domain-tuned networks, the results reveal key signs of transferable spatial reasoning and cross-task adaptability. The model shows an ability to localize anatomical structures, preserve fine textures, and recover structural continuity—without any medical-domain training. These findings highlight that even without explicit optimization, video models inherently learn temporal and structural priors that align with medical imaging needs, indicating their strong potential as a universal visual backbone for clinical applications.

% \smallskip
% \noindent\textbf{Broader validation and adaptation.}
% Our experiments focus on thoracic and abdominal CT data, where respiratory motion dominates. Broader validation across diverse anatomical regions and modalities (e.g., MRI, ultrasound, and PET) will be essential to evaluate robustness and generalization. Future work should explore lightweight domain adaptation strategies or instruction-based prompting to improve precision while maintaining the zero-shot flexibility of foundation video models.

\smallskip
\noindent\textbf{Toward a unified medical foundation model.}
Ultimately, these efforts aim toward a unified medical foundation model built upon large video architectures—capable of performing segmentation, enhancement, motion prediction, and temporal reasoning across diverse clinical tasks and modalities, all without task-specific retraining. Such a model would represent a transformative step toward general-purpose, adaptive medical AI that can learn, reason, and generalize across space and time.

\section{Conclusion}

In this work, we explore the zero-shot capabilities of large video models in the medical imaging domain. Using radiotherapy organ motion prediction as a challenging testbed, we show that a general-purpose video model—without any medical-domain training—can learn and reason about patient-specific respiratory motion directly from sequential CT data. The model produces temporally coherent and anatomically consistent predictions, surpassing specialized pipelines that rely on deformation-field supervision or task-specific architectures.

Beyond motion prediction, we demonstrate that the same model can perform segmentation, denoising, and super-resolution in a zero-shot manner. Although its performance does not yet match fully supervised approaches, these results clearly reveal inherent adaptability and cross-task generalization of large video models. Such capabilities mark significant step toward unified visual model in medicine.

Looking forward, we envision extending this line of research toward developing medical foundation models built upon large video architectures—capable of performing diverse imaging tasks with minimal supervision. This paradigm promises scalable, adaptive, and clinically relevant AI systems that can learn, interpret, and reason across time, anatomy, and modality.

\section*{Acknowledgments}
This research is supported in part by the National Institutes of Health under Award Numbers R01EB032680, R01CA272991, R01DE033512, and U54CA274513.
% \section{Acknowledgement}

\clearpage

{
    \small
    \bibliographystyle{ieeenat_fullname}
    \bibliography{ref, zzhou}
}

% WARNING: do not forget to delete the supplementary pages from your submission 
% \input{sec/X_suppl}

\end{document}